\documentclass{article}

\usepackage[final]{neurips_2024_calm}

\usepackage[utf8]{inputenc} 
\usepackage[T1]{fontenc}    
\usepackage{hyperref}       
\usepackage{url}            
\usepackage{booktabs}       
\usepackage{amsfonts}       
\usepackage{nicefrac}       
\usepackage{microtype}      
\usepackage{xcolor}         

\usepackage{graphicx}               
\usepackage{multirow}               
\usepackage{subcaption}             
\usepackage{amsmath}                
\usepackage{amssymb}                
\usepackage[capitalise]{cleveref}   

\newtheorem{definition}{Definition}

\DeclareRobustCommand{\parhead}[1]{\textbf{#1}~}

\newcommand{\rmdo}{\mathrm{do}}
\newcommand{\ie}{\mathrm{IE}}
\newcommand{\tubingen}{T{\"u}bingen }

\newcommand{\mb}[1]{\mathbf{#1}}

\title{Evaluating Interventional Reasoning Capabilities of Large Language Models}

\author{Tejas Kasetty\thanks{Correspondence to \texttt{tejas.kasetty@mila.quebec} } $\;^{1, 4}$
\And
Divyat Mahajan$^{1, 4}$ 
\And
Gintare Karolina Dziugaite$^{2, 4}$
\AND
Alexandre Drouin$^{3, 4}$ 
\hspace{8ex}
Dhanya Sridhar$^{1, 4}$ \\
\AND
\textnormal{$^1$ Université de Montréal}
\And
\textnormal{$^2$ Google DeepMind}
\And
\textnormal{$^3$ ServiceNow Research}
\And
\textnormal{$^4$ Mila}
}

\begin{document}

\maketitle

\begin{abstract}
  Numerous decision-making tasks require estimating causal effects under interventions on different parts of a system. As practitioners consider using large language models (LLMs) to automate decisions, studying their causal reasoning capabilities becomes crucial. A recent line of work evaluates LLMs ability to retrieve commonsense causal facts, but these evaluations do not sufficiently assess how LLMs reason about interventions. Motivated by the role that interventions play in causal inference, in this paper, we conduct empirical analyses to evaluate whether LLMs can accurately update their knowledge of a data-generating process in response to an intervention. We create benchmarks that span diverse causal graphs (e.g., confounding, mediation) and variable types, and enable a study of intervention-based reasoning. These benchmarks allow us to isolate the ability of LLMs to accurately predict changes resulting from their ability to memorize facts or find other shortcuts. We evaluate six LLMs on the benchmarks, finding that GPT models show promising accuracy at predicting the intervention effects.
\end{abstract}

\section{Introduction}
\label{sec:intro}

Large language models (LLMs) have achieved impressive performance on a variety of human-relevant tasks, from summarizing web-based information and answering complex questions, to carrying out tasks as web-based agents~\citep{chen2021evaluating, brown2020language, li2022competition, katz2024gpt,drouinworkarena}. As LLMs increasingly become used to make decisions, which fundamentally require understanding the causal impact various actions can have, there has been a recent push to evaluate whether LLMs demonstrate causal reasoning ability \citep{cai2023knowledge,jin2023can,jin2024cladder,kiciman2024causal,liu2024llms}.
The challenge for this emerging field is to operationalize notions of causal reasoning that can be verbalized as text-based questions for LLMs. 
\citet{kiciman2024causal} address this question by evaluating the ability of LLMs to retrieve known cause-and-effect relations.
In contrast, \citet{jin2023can,jin2024cladder,liu2024llms,cai2023knowledge} focus on defining various queries that could only be solved with knowledge of causality and graphical models, testing abstract causal reasoning that could generalize to unseen contexts.
In this paper, we contribute to the growing body of work that evaluates the causal reasoning capabilities of LLMs by focusing on an aspect of causality that is both intuitive for humans \citep{waldmann2005seeing} and crucial for decision-making \citep{Binz2023}: the ability to adapt our beliefs in response to \emph{interventions}.

The notion of intervening on a variable is at the core of causality. In this paper, we focus on perfect interventions, where a variable in a system is manipulated and set to a particular new value. Interventions modify the causal graphical model \citep{pearl2009causality} of a system by deleting all incoming edges to the intervened variable. Experiments in the field of cognitive psychology \citep{waldmann2005seeing} suggest that humans instinctively recognize that an action (i.e., an intervention) changes existing causal relationships, correctly making different inferences before and after interventions are performed on various variables.  

Besides being natural for humans, reasoning about interventions is crucial for causal inference. 
Consider the example of an LLM agent that acts as a ``science assistant.'' It knows about some causal relationships and must infer new ones given observational evidence to suggest candidate experiments. For ease, consider three variables, $A$, $B$ and $C$, where $A$ is known to cause $B$ and $C$, but where the effect that $B$ has on $C$ is unknown. Initially, the agent cannot conclude anything about the effect of $B$ on $C$ due to the confounding effect of $A$, but if it encounters new information that $B$ was intervened on, and $B$ was observed to be correlated with $C$, the agent should correctly conclude that the intervention severs the confounding effect of $A$, allowing it to infer that $B$ must actually have a causal effect on $C$.
The ability to understand how interventions affect known causal relationships is central to drawing new causal conclusions when presented with new evidence.

The key contribution of this paper is to evaluate intervention reasoning in LLMs by introducing \emph{intervention effects}, binary classification tasks that evaluate which causal relations in a graph are modified by an observed intervention. We provide a framework for verbalizing a wide range of intervention effects to LLMs, varying the choice of causal graphs and names of variables. With this framework, we develop a benchmark of intervention effect tasks that help us disentangle the intervention reasoning capabilities of LLMs from other factors that contribute to performance such as memorization of facts and the ability to extract graphs from text. 

\section{Related Works} This paper relates most closely to recent papers that develop benchmarks to evaluate LLMs on various causal reasoning tasks. 
\citet{kiciman2024causal} introduced multiple causal reasoning benchmarks for LLMs, including evaluating the ability of LLMs to recover the bivariate causal DAGs introduced in the T{\"u}bingen pairs dataset \citep{mooij2016distinguishing}. \citet{kiciman2024causal} found that GPT models recovered known causal relationships with up to 96\% accuracy when experimenting with various prompting strategies such as including system prompts.
However, evaluating LLMs on their ability to retrieve causal knowledge about known variables constitutes \emph{commonsense causal reasoning}.
In contrast, this paper contributes to work that evaluates abstract causal reasoning \citep{Binz2023,jin2024cladder,jin2023can,cai2023knowledge,liu2024llms}, assessing the ability of LLMs to use axioms of causality to solve tasks involving general or even new variables.

In the vein of causal reasoning, \citet{jin2023can} studied whether LLMs could correctly infer some causal relationships based on conditional independence statements, comparing LLM predictions to those made by an oracle causal discovery algorithm for observational data, where all causal relations cannot be resolved. 
In contrast to observational causal discovery, this paper focuses on reasoning about interventions.
Also focusing on interventions, \citet{jin2024cladder} introduced CLadder, a comprehensive benchmark that includes the estimation of causal effects from quantitative data. Causal effect estimation is a complex task that requires solving multiple sub-tasks such as: (i) parsing the prompt to extract a causal DAG, (ii) inferring a function that estimates the effect given the DAG, and (iii) applying that function to the given quantitative data.
Concurrently, \citet{cai2023knowledge} introduced a task that asks LLMs to output only causal relationships given a tabular dataset that includes variable names. They focus on disentangling the impacts that prior knowledge (e.g., variables names) and quantitative data have on LLM performance.
The empirical study we conduct to assess whether LLMs are sensitive to the presence of plausibly memorized causal relations is similar to experiments conducted by \citet{cai2023knowledge}.
In contrast to these benchmarks, intervention effects target a narrower question than the general estimation of causal effects, since intervention effects involve binary classification only (i.e., the absence/presence of causal relations in DAGs). 
We argue that the evaluations we design better isolate causal reasoning from sub-tasks like drawing statistical inferences from quantitative data provided in-context, which  CLadder and \citet{cai2023knowledge} require.

In focusing on intervention effects, we build on the work of \citet{Binz2023}, who were motivated by prior work in psychology \citep{waldmann2005seeing} that shows humans weight collected observational evidence and experimental evidence differently when drawing causal conclusions.
\citet{Binz2023} adapted this psychology study for LLMs, creating prompts that describe observational and post-interventional findings to LLMs to see if they update their beliefs about a system after interventions. They found that GPT-3, unlike human subjects, fared poorly at understanding the implications of interventions.
Motivated by their focus on intervention-based reasoning, we significantly expand on the evaluations designed by \citet{Binz2023}, systematically generating intervention effects with varying degrees of difficulty to further explore the effects of plausible memorization and shortcuts like relation retrieval.



\section{Understanding changes in Causal Relationships via Intervention Effects }
\label{sec:prelim}

We begin by summarizing causal directed acyclic graphical models (DAGs) and perfect interventions, the two concepts central to this work. After reviewing these concepts, we introduce intervention effects, the binary prediction tasks that serve as key contributions of this paper.

\subsection{Background}
A causal DAG $G = (V=\{V_1, \ldots, V_n\}, E)$ is defined by a vertex set $V$ that consists of random variables $\{V_1, \ldots, V_n\}$ and directed edges in the set $E$ that represent causal relationships between the variables. A DAG defines a joint distribution over the random variables as:
\begin{align}
    P(V_1, \ldots, V_n) = \prod_{i=1}^n P(V_i \vert \mathrm{Pa}^G_i),
\end{align}
where $\mathrm{Pa}^G_i$ denotes the parents of $V_i$ in the DAG $G$. In general, we say that a causal relation exists between two variables $u$ and $v$ when there is a path from $u$ to $v$ in $G$.
 
\Cref{fig:graphs} in the appendix illustrates the three causal DAGs that we focus on in this paper: bivariate, confounding, and mediation. In the confounding graph, the variable $A$ causes both $B$ and $C$, thus confounding our ability to infer the causal effect that $B$ has on $C$ from observational data alone. In the mediation graph, the variable $A$ only has an indirect effect on $C$ via $B$.

Causal DAGs also entail distributions after interventions to these variables, distinguishing them from standard DAG models. In this paper, we focus on a class of interventions called \emph{perfect interventions}, represented using the operation $\rmdo(V_i=v)$, which means that the variable $V_i$ is set to the value $v$. We define the distribution over variables post-intervention by modifying $G$ to delete all incoming edges to $V_i$, severing the links between $V_i$ and its parents.

\subsection{Intervention effects}
To formalize intervention effects, the key binary classification task that we consider in this work, we begin by defining causal relations in DAGs precisely.
\begin{definition}[Causal relation]
Given a causal DAG $G=(V,E)$, we say that there exists a causal relation between $V_u$ and $V_v$ in DAG $G$ if there exists a directed path $e_{ua} \rightarrow \ldots \rightarrow e_{bv}$ in $G$ from $V_u$ to $V_v$ (where each edge $e_{ij}$ along the path is in $G$). We define an indicator variable $C_{uv}(G) \in \{0, 1\}$, such that $C_{uv}(G) = 1$ if and only if there is a causal relation between $V_v$ and $V_u$.
\end{definition}
In words, a causal relation captures whether or not a variable exerts an indirect or direct causal influence on another variable in a particular DAG $G$ that contains these variables. 

\begin{definition}[Intervention effect]
Given a causal DAG $G=(V,E)$, a variable $V_i$ on which we perform a perfect intervention captured by $\rmdo(V_i = *)$ (where we use ``*'' to indicate that we do not care about the value that $V_i$ is set to), and a query causal relation $C_{uv}$, an \textbf{intervention effect} (IE) defines a binary classification task as follows,
\begin{align}
    \ie^G_i(C_{uv}) = C_{uv}(G)- C_{uv}(G^i), \label{eq:intervention-effect}
\end{align}
where the DAG $G^i$ is the modification of DAG $G$ under an intervention to the variable $V_i$. 
\end{definition}
An $\ie^G_i(C_{uv})$ is defined with respect to a DAG $G$ and intervention target $V_i$, and assigns each causal relation $C_{uv}$ to a binary label of 1 or 0. When a causal relation $C_{uv}$ in $G$ is modified by a perfect intervention on $V_i$, i.e., the relation $C_{uv}$ is different in the modified DAG $G^i$, $\ie^G_i(C_{uv})$ is 1 (and 0 otherwise).
Intuitively, since interventions sever edges between a variable and its parents, $\ie^G_i(C_{uv})$ is 1 when the intervention target is a variable along the path from $u$ to $v$ (including $v$ itself). 

Given a causal DAG $G$, each possible intervention that we can perform defines a different classification task over causal relations, defined by a particular $\ie^G_i(\cdot)$, which serves as a labeling function for that task. 
We focus on the three causal DAGs in \Cref{fig:graphs} since they sufficiently capture many real scenarios without introducing unnecessary complexity that could render the empirical findings ambiguous.
The next section focuses on how we evaluate LLMs on these IE classification tasks in a zero-shot way (i.e., without training examples) by verbalizing the tasks as prompts.

\section{Evaluating LLMs on Intervention Effects}
\label{sec:method}

\begin{figure*}[t]
    \centering
    \vspace{-1cm}
    \includegraphics[width=0.70\textwidth]{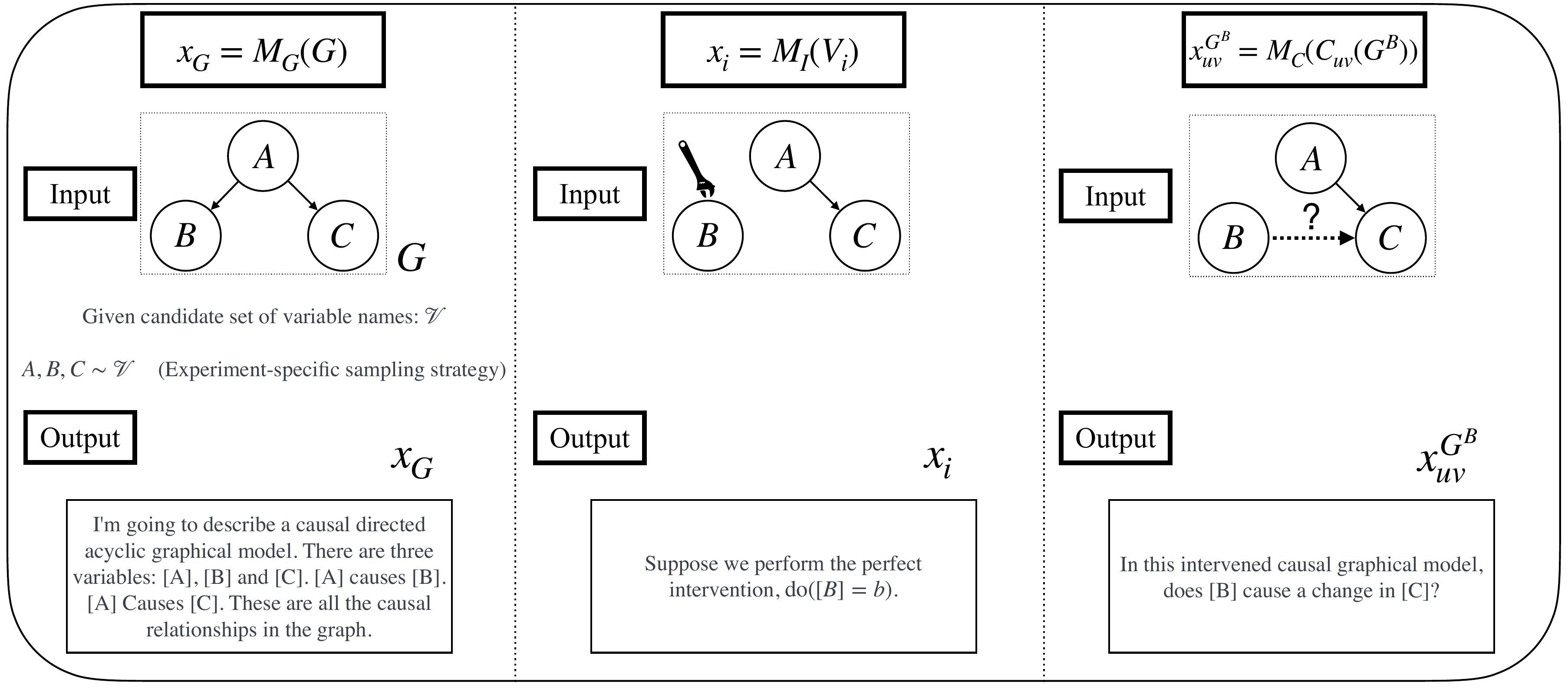}
    \caption{An illustration of the mapping functions that verbalize the information for a single intervention effect estimation task into a prompt.}
    \label{fig:prompt}
\end{figure*}

To evaluate an LLM on a binary classification problem defined by $\ie^G_i(C_{uv})$, we need to verbalize the causal DAG $G$, the concept of an intervention as well as the intervention target $V_i$, and the causal relation $C_{uv}$ of interest.
\Cref{fig:prompt} illustrates how we verbalize these three steps to generate prompt.

\parhead{Step 1: Generating variable names.} To verbalize an input causal DAG $G=(V,E)$, we select names for the variables $V$ and describe each edge $E_{ij} \in E$ using the format ``[$i$] causes [$j$].'' We further specify that these are all known causal relationships in the graph to avoid ambiguities, e.g., around the presence of unobserved confounding variables. In the empirical studies, we assess multiple ways of choosing variable names, designing studies to tease apart how well LLMs generalize rather than relying on variable names and facts that were potentially encountered during training.

\parhead{Step 2: Describing an intervention.} In the second step, we specify that a perfect intervention is performed, during the notation $\rmdo(V_i=v)$ since the $\rmdo$-operator could have been encountered during training. To further study the robustness of LLMs to memorizing facts from training, the empirical studies include experiments that use random strings to describe $\rmdo$-operator.

\parhead{Step 3: Verbalizing the binary classification problem.} In the final step, we verbalize the target $\ie^G_i(C_{uv})$, using the phrase ``does [$u$] cause a change in [$v$]?'' to query the presence or absence of a causal relation between $u$ and $v$. 



\parhead{Evaluation metric.} After verbalizing an intervention effect as a prompt to an LLM, we receive a yes/no response that we refer to as $\hat{\ie}^G_i(C_{uv})$.
Although prediction accuracy is a natural evaluation metric in this setting, it can misleading in this setting. To see this, we begin by noting that the LLM forms some belief about the presence or absence of a causal relation in a DAG $G$, denoted $\hat{C}_{uv}(G)$. Consider the scenario where,
\begin{align}
\nonumber
\begin{split}
     &C_{uv}(G) = 1, \,\, C_{uv}(G^i) = 1, \,\, \\
     &\hat{C}_{uv}(G) = 0, \,\, \hat{C}_{uv}(G^i) = 0.
\end{split}
\end{align}
In this example, a target causal relation $C_{uv}$ is true in both the base causal DAG $G$ and its modified, post-intervention counterpart $G^i$, but the LLM incorrectly predicts that these causal relations are false under both graphical scenarios. The accuracy metric misleads us when the LLM correctly predicts that the causal relation does not vary, but does not parse the causal relation correctly in either graphical scenario. Thus, to ensure that the accuracy is 0 in such cases, we slightly modify the accuracy metric to be,
\begin{align}
    \label{eq:acc_metric}
     \mathbf{1}[\ie^G_i(C_{uv}) =  \hat{\ie}^G_i(C_{uv})] \cdot \mathbf{1}[C_{uv}(G) = \hat{C}_{uv}(G)]
\end{align}
Now, an LLM's prediction is only considered correct when it understands the DAG $G$ correctly and accurately solves an IE problem. 
To infer the belief $\hat{C}_{uv}(G)$ that an LLM has about a causal relation, we make a minor modification to the prompt in \Cref{fig:prompt}, omitting the second step that verbalizes an intervention, and asking the LLM about a causal relation but in the ``observed causal graphical model'' instead of the intervened one. These additional prompts allow us to evaluate LLMs on IEs.

In the next section, we investigate several research questions about LLMs and intervention reasoning ability using the proposed framework to define a suite of IE problems.

\section{Empirical Analysis}
\label{sec:experiment}
To study the ability of LLMs to understand interventions, we use the introduced framework to develop three benchmarks that differ in how variable names are selected to verbalize IEs: 
\begin{enumerate}
	\item{\bf Random Char}: Variable names are randomly chosen English characters. 
	\item{\bf \tubingen}: Variables names are chosen from entities that appear in the \tubingen pairs (TP) dataset of causal relations~\citep{mooij2016distinguishing} such that some relations occur in TP.
	\item{\bf Random \tubingen}: Variable names are chosen from entities that appear in the TP dataset so that no causal relations exist in the TP data.
\end{enumerate} 
See \Cref{app:dataset} for an illustration of the full details about the datasets.
For each benchmark and IE, we sample variable names fifteen times to report significant differences.

We investigate four research questions (RQs) on the proposed benchmarks, studying four LLMs: GPT-3.5, GPT-4, GPT-4-turbo~\citep{openai2023gpt}, and LLaMA-2~\citep{touvron2023llama}.
Unless otherwise specified, for each IE, we aggregate results over the enumerated causal relations.
We present the findings for the first two questions in the main paper and discuss the remaining in  \Cref{app:extra-results}


\parhead{RQ1: How accurate are at LLMs at IEs generated with random characters as variables?} To study this first question, we evaluate the performance of the four LLMs on the Random benchmark.
\Cref{tab:ie_random} summarize these results. We see that GPT-based models perform notably better than the LLaMA models, with GPT-4-turbo demonstrating near-perfect accuracy across all effects. LLaMA-2 and LLaMA-3's performance suggests that they do not reliably model interventions. In what follows in the main paper, we focus on results for GPT models, deferring LLaMA results to \Cref{app:extra-res-rq1}.

\begin{table}
  \caption{\textbf{IE prediction accuracy on the Random Char benchmark}. GPT-4 variants are the best performing models, while LLaMA-2 appears to struggle with interventional reasoning. Top performaning model is indicated by bolded figures.}
  \label{tab:ie_random}
  \centering
  \resizebox{\textwidth}{!}{
    \begin{tabular}{ccccccccc}
        \toprule
        Graph Type &  \multicolumn{2}{c}{\bf Bivariate} &  \multicolumn{3}{c}{\bf Confounding} &  \multicolumn{3}{c}{\bf  Mediation} \\
        \midrule
        Intervened Variable & A & B & A & B & C & A & B & C \\
        \midrule
        GPT-3.5 & $0.83 \pm 0.08$ & $0.87 \pm 0.06$ & $0.80 \pm 0.09$ &  $0.69 \pm 0.12$ &  $0.36 \pm 0.09$ &  $0.58 \pm 0.11$ &  $0.36 \pm 0.12$ &  $0.67 \pm 0.12$ \\
        \cmidrule{1-9}
        GPT-4 & $1.0 \pm 0.0$ & $1.0 \pm 0.0$ & $1.0 \pm 0.0$ & $1.0 \pm 0.0$ & $1.0 \pm 0.0$ & $0.78 \pm 0.09$ & $0.82 \pm 0.08$ & $0.96 \pm 0.03$ \\
        \cmidrule{1-9}
        \bf{GPT-4-turbo} & $\mb{1.0 \pm 0.0}$ & $\mb{0.97 \pm 0.03}$ & $\mb{0.96 \pm 0.03}$ & $\mb{0.93 \pm 0.05}$ & $\mb{1.0 \pm 0.0}$ & $\mb{0.98 \pm 0.02}$ & $\mb{1.0 \pm 0.0}$ & $\mb{1.0 \pm 0.0}$ \\
        \cmidrule{1-9}
        GPT-4o & $1.0 \pm 0.0$ & $1.0 \pm 0.0$ & $0.98 \pm 0.02$ & $0.91 \pm 0.04$ & $1.0 \pm 0.0$ & $0.93 \pm 0.05$ & $1.0 \pm 0.0$ & $0.87 \pm 0.06$ \\
        \cmidrule{1-9}
        LLaMA-2 & $0.50 \pm 0.12$ & $0.40 \pm 0.12$ & $0.56 \pm 0.12$ &  $0.53 \pm 0.11$ &  $0.16 \pm 0.06$ &  $0.69 \pm 0.09$ & $0.56 \pm 0.12$ & $0.64 \pm 0.12$ \\
        \cmidrule{1-9}
        LLaMA-3 & $0.80 \pm 0.09$ & $0.83 \pm 0.06$ &  $0.51 \pm 0.1$ &  $0.47 \pm 0.08$ & $0.96 \pm 0.03$ &  $0.38 \pm 0.1$ &  $0.53 \pm 0.08$ &  $0.4 \pm 0.1$ \\
        \bottomrule
    \end{tabular}}
\end{table}

\parhead{RQ2: To what extent is LLM performance affected by possibly memorized causal relations?}
While we might be tempted to conclude that GPT-4 reliably predicts changes to models after interventions are performed, we consider spurious factors that can affect model performance.
In particular, \citet{kiciman2024causal} found that GPT models reliably retrieved information about TP causal relations, suggesting that these relations could have been included in the training data for LLMs. (We reproduce their findings in Appendix \Cref{tab:amit_tubingen}.)
This leads to a worrying possibility: after interventions, could LLMs fail to update their beliefs about causal relations that they have potentially memorized? 
To study this, we consider only the subset of causal relations that appear in TP and the interventions that \emph{sever these relations}. If an LLM achieves good performance in general by having memorized some known causal relations, then it would achieve worse performance on \tubingen for the selected IEs, which sever known causal relations and go against the LLM's learned biases, compared its performance on Random or Random \tubingen, where post-interventional causal relations do not directly contradict known facts.
In \Cref{app:extra-res-rq2}, we specify which causal relations appear in TP for each causal DAG we study, and which interventions modify these relations. 
\Cref{tab:ie_memorization} summarizes the results. Bolded figures indicate that the performance on \tubingen drops significantly (with $\alpha=0.05$) compared to performance on either of the other benchmarks. Interestingly, GPT do not show evidence of relying purely on memorized causal relations, since performance is overall comparable across the benchmarks. Further, the models seem to struggle with the Random \tubingen benchmark more than they do with the other benchmarks, leading to a question about whether they are in fact sensitive to variations on the TP dataset.

\begin{table}
  \caption{\textbf{IE prediction accuracy considering only causal relationships that the LLM could potentially memorize.} For causal relations that appear in TP, and for interventions that modify these relations, worse performance on the \tubingen benchmark compared to performance on Random Char or Random \tubingen provides evidence that LLMs might be relying on memorized facts to achieve high accuracy on IEs. Bolded figures indicate performances that are significantly ($\alpha=0.05$) worse than the corresponding performance on \tubingen.}
  \label{tab:ie_memorization}
  \centering
  \resizebox{0.9\textwidth}{!}{
    \begin{tabular}{cccccc}
        \toprule 
        \textbf{Model} & \textbf{Graph} & \textbf{Intervened Variable} & \textbf{Random Char} & \textbf{\tubingen}& \textbf{Random \tubingen}  \\
        \midrule
        \multirow[c]{4}{*}{GPT-3.5} & \multirow[c]{1}{*}{Bivariate} & B  & $0.8 \pm 0.1$ & $0.8 \pm 0.1$ & $0.53 \pm 0.13$ \\
    \cmidrule{2-6}
     & \multirow[c]{1}{*}{Confounding} & C & $0.47 \pm 0.13$ & $0.4 \pm 0.13$ & $0.47 \pm 0.13$ \\
     \cmidrule{2-6}
    & \multirow[c]{2}{*}{Mediation} & B & $0.4 \pm 0.13$ & $0.4 \pm 0.13$ & $0.4 \pm 0.13$ \\ 
    &  & C & $0.67 \pm 0.12$ & $0.67 \pm 0.12$ & $0.6 \pm 0.13$ \\
    \cmidrule{1-6} 
    \multirow[c]{4}{*}{GPT-4} & \multirow[c]{1}{*}{Bivariate}  & B & $1.0 \pm 0.0$ & $1.0 \pm 0.0$ & $1.0 \pm 0.0$ \\
    \cmidrule{2-6}
    & \multirow[c]{1}{*}{Confounding} & C & $1.0 \pm 0.0$ & $1.0 \pm 0.0$ & $1.0 \pm 0.0$ \\
    \cmidrule{2-6} & \multirow[c]{2}{*}{Mediation} & B & $0.8 \pm 0.1$ & $\bf{0.73 \pm 0.11}$ & $0.93 \pm 0.06$ \\
    &  & C & $0.93 \pm 0.06$ & $1.0 \pm 0.0$ & $0.93 \pm 0.06$ \\
    \cmidrule{1-6}
    \multirow[c]{4}{*}{GPT-4-turbo} & \multirow[c]{1}{*}{Bivariate} & B  & $0.93 \pm 0.06$ & $0.8 \pm 0.1$ & $0.73 \pm 0.11$ \\
    \cmidrule{2-6}
    & \multirow[c]{1}{*}{Confounding} & C & $1.0 \pm 0.0$ & $1.0 \pm 0.0$ & $1.0 \pm 0.0$ \\
    \cmidrule{2-6} & \multirow[c]{2}{*}{Mediation}  & B & $1.0 \pm 0.0$ & $1.0 \pm 0.0$ & $0.93 \pm 0.06$ \\
    &  & C & $1.0 \pm 0.0$ & $1.0 \pm 0.0$ & $0.93 \pm 0.06$ \\
    \cmidrule{1-6}
    \multirow[c]{4}{*}{GPT-4o} & \multirow[c]{1}{*}{Bivariate} & B  & $1.0 \pm 0.0$ & $0.93 \pm 0.06$ & $0.73 \pm 0.11$ \\
    \cmidrule{2-6}
    & \multirow[c]{1}{*}{Confounding} & C & $1.0 \pm 0.0$ & $1.0 \pm 0.0$ & $1.0 \pm 0.0$ \\
    \cmidrule{2-6} & \multirow[c]{2}{*}{Mediation}  & B & $1.0 \pm 0.0$ & $1.0 \pm 0.0$ & $0.87 \pm 0.09$ \\
    &  & C & $0.93 \pm 0.06$ & $0.87 \pm 0.09$ & $0.8 \pm 0.1$ \\
    
    \bottomrule
    \end{tabular}}
\end{table}

\section {Discussion and Limitations}

The goal of this paper is to introduce a causal reasoning benchmark that stress-tests the ability of LLMs to accurately predict how knowledge should be updated after interventions are performed, without conflating other aspects of reasoning such as statistical inference on quantitative data.
Our findings are optimistic, but we believe that nevertheless, they underscore the continued need for benchmarks and studies that evaluate varied aspects of abstract causal reasoning in LLMs, especially if practitioners wish to use LLMs to generate candidate decisions. 

While the intervention effect prediction task we define in this paper has the benefit of being easy to evaluate, since it requires binary responses, the findings that this task can suggest are also limited. For example, IE prediction cannot help us assess how accurately LLMs perform \emph{causal identification}, the process of deciding which causal inferences can be made given a causal DAG. Moreover, we focus on evaluation in this paper and do not propose methods for improving causal reasoning in LLMs via few-shot learning or fine-tuning. Both of these limitations point to future research directions that we think are worth exploring.


\section*{Acknowledgements}

We would like to thank Tejas Vaidhya for contributing to experiments during the initial phase of the project. 
DS acknowledges support from NSERC Discovery Grant RGPIN-2023-04869, and a Canada-CIFAR AI Chair. 

{\small
\setcitestyle{numbers}
\bibliographystyle{plainnat}
\bibliography{references}}

\newpage 

\appendix

\section{Experiment Setup Details}
\label{app:full-details}

\subsection{Causal DAGs}

\begin{figure*}[h!]
    \centering
    \includegraphics[width=.8\textwidth]{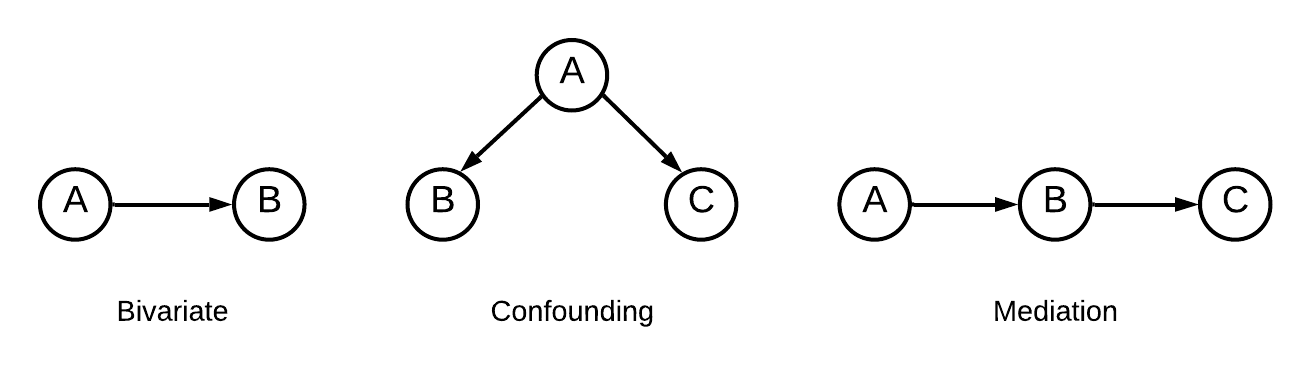}
    \caption{\textbf{Causal DAGs.} In the empirical studies, we define intervention effects based on three causal DAGs: bivariate, confounding and mediation graphs.}
    \label{fig:graphs}
\end{figure*}

We consider 3 basic causal DAGs: Bivariate, Confounding and Mediation as shown in \Cref{fig:graphs}

\subsection{Dataset Generation}
\label{app:dataset}

\paragraph{Random Char (R).} In the causal graphs used in this benchmark, the nodes are labeled with arbitrary English letters (e.g. 'z', 'b') that are independently generated for each graph. 

\paragraph{\tubingen (T).} For the \tubingen benchmark, we select the graph nodes from the \tubingen pairs (TP) dataset in the following manner:
\begin{itemize}
    \item {\textbf{Bivariate.} We sample cause-effect pairs from the TP dataset and each cause-effect pair is used to generate a graph. \\
    Example - Altitude (A) - Temperature (B) is a cause-effect pair in the TP dataset, and the input graph states:  \textit{Altitude causes temperature}.}
    \item {\textbf{Confounding.} For variable $B$ and $C$, we first  sample cause-effect pairs and then for sampled pair we randomly select a variable (for $A$) from the TP dataset that it is different from the variables in the corresponding cause-effect pair. \\
    Example - Altitude ($B$) - Temperature ($C$) is the sampled cause-effect pair and Age ($A$) is the randomly selected variable from the TP dataset. The relationships in the input graph are as follows: 
    \textit{Age causes altitude ; Age causes temperature}.}
    \item {\textbf{Mediation.} For variable $A$ and $C$, we first randomly sample cause effect pairs and then for sampled pair we randomly select a variable (for $B$) from the TP dataset that it is different from the variables in the corresponding cause-effect pair. \\ 
    Example - Altitude ($A$) - Temperature ($C$) is the sampled cause-effect pair and Age (B) is the randomly selected variable from the TP dataset.  The relationships in the input graph are as follows: 
    \textit{Altitude causes age ; Age causes temperature}.}
\end{itemize}
The rationale behind defining the \tubingen dataset in this manner is elaborated in \Cref{app:extra-res-rq2}.

\paragraph{Random \tubingen (RT).} Similar to the \tubingen case,  we select the graph nodes from the set of variables present in the TP dataset. However, none of the causal relations defined in the graphs are present in the TP dataset. Instead the causal relations defined are between randomly selected variables, as follows:
\begin{itemize}
    \item {\textbf{Bivariate.} We randomly sample two unrelated variables from the TP dataset and define a cause-effect relationship in the input graph. \\
    Example - Age ($A$) - Temperature ($B$) are variables without a cause-effect relation in the TP dataset, and the input graph states:  \textit{Age causes Temperature}.}
    \item {\textbf{Confounding.} We randomly sample 3 variables from the TP dataset that no two variables selected have a causal relation in the TP dataset and we define the graph. \\ 
    Example - Altitude (A) - Horsepower (B) - Cement (C) are randomly selected variables from the TP dataset. No two variables have a causal relationship in the TP dataset. The relationships in the input graph is as follows: 
    \textit{Altitude causes Horsepower ; Altitude causes Cement}.}
    \item {\textbf{Mediation.}  We select the 3 variables exactly as we did for the confounding case and define the graph. \\ 
    Example - Altitude (A) - Horsepower (B) - Cement (C) are randomly selected variables from the TP dataset. The relationsihips in the input graph is as follows:
    \textit{Altitude causes Horsepower ; Horspower causes Cement}.}
\end{itemize}

\subsection{Details regarding LLMs}

\subsubsection{Querying LLMs.}
To query the GPT models, we used the OpenAI API \footnote{\url{https://openai.com/blog/openai-api}} through the Langchain  interface \footnote{\url{https://python.langchain.com/docs/expression_language/interface}}. Meanwhile, VLLM library \footnote{\url{hhttps://docs.vllm.ai/en/latest/}} was used for fast inference from LLaMA models.

Regarding compute resources, since the GPT models are hosted remotely by OpenAI, we were able to make API calls locally with minimal CPU usage. Conversely, for LLaMA models, we first had to load the model onto a cluster equipped with a GPU (A100/80 GB RAM) before submitting our queries. 

\subsubsection{LLM Output.}
To ensure that LLMs respond with a yes or no to queries, we formalized a response format requiring LLMs to encapsulate their yes or no answers within an \texttt{<answer></answer>} tag. If the LLM does not adhere to this format, we initiate a retry, instructing it to comply with the required format. After the first retry, we relax the response format requirements, expecting only a yes or no answer. If after 10 retries the LLM fails to meet this criterion, we mark the attempt as a failure and attribute zero accuracy on the intervention effect prediction task. 

\subsubsection{Model Ids.}
The specific model-ids of the LLMs are:
\begin{itemize}
    \item GPT-3.5: \texttt{gpt-3-turbo-16k}
    \item GPT-4: \texttt{gpt-4}
    \item GPT-4-turbo: \texttt{gpt-4-turbo}
    \item GPT-4o: \texttt{gpt-4o}
    \item LLaMA-2: \texttt{llama-2-7b}
    \item LLaMA-3: \texttt{llama-3.1-8b}
\end{itemize}

\subsection{Substitution Task}

\begin{figure*}[t]
    \centering
        \includegraphics[width=0.9\textwidth]{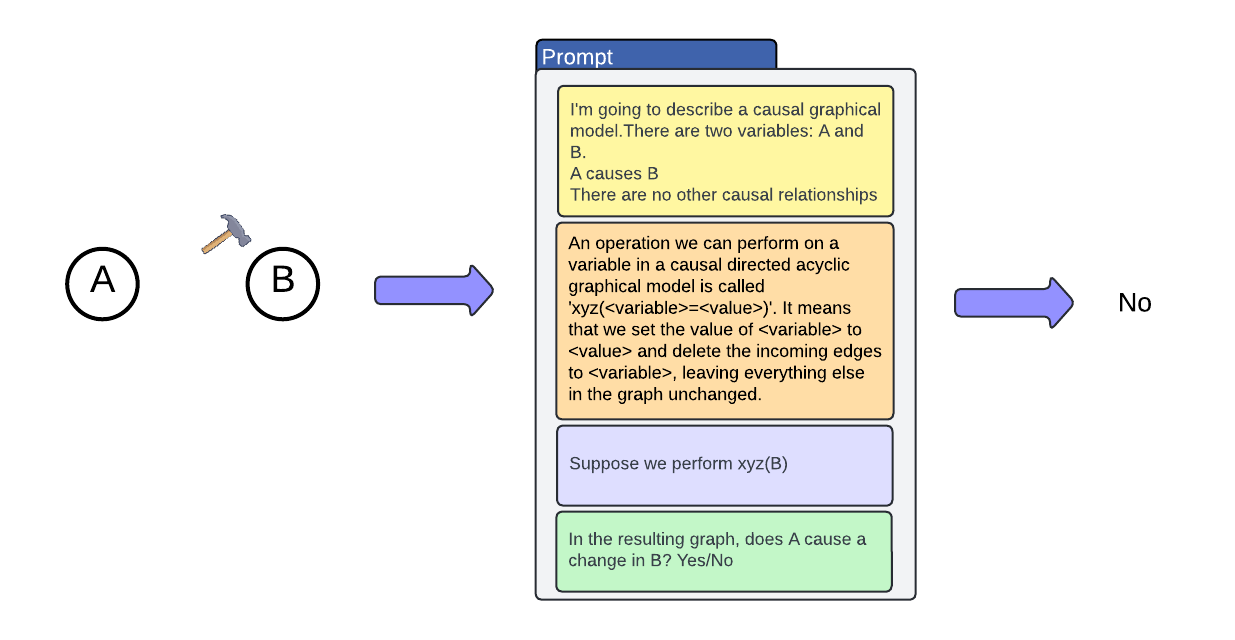}
        \caption{\textbf{Prompt design for intervention reasoning in the substitution task.} Instead of using the word intervention directly in the prompt, we illustrate the concept of interventions and define it with a random word, for instance,  operation 'xyz' in the prompt template above.}
    \label{fig:prompt-substitution}
\end{figure*}

For this task, we first describe the concept of interventions to LLM and refer to them as some arbitrary string of characters (E.g. 'xyz'). We then follow the same prompt template (\Cref{fig:prompt}) to test the intervention effect performance, except we never use the word intervention again and replace it with the chosen substitution word. The prompt verbalization for this task is illustrated in \Cref{fig:prompt-substitution}.

\subsection{Ground Truth Intervention Effects}
\label{app:true-ie}

\begin{table}[t]
    \caption{Ground truth intervention effects for all the scenarios (causal graph, intervention variable, causal relation) considered in our benchmark.}
    \label{tab:true_ie}
    \centering
    \begin{tabular}{ccccc}
            \toprule
            \multicolumn{1}{c}{\bf Graph}  &\multicolumn{1}{c}{\bf Intervention} &\multicolumn{1}{c}{\bf Questions} &\multicolumn{1}{c}{\bf Ground Truth IE} \\
            \midrule
            \multirow{4}{*}{Bivariate}  
               & \multirow{2}{*}{A} & $A \to B$ & 0\\ 
                                  & & $B \to A$ & 0 \\
                \cmidrule{2-4}
               & \multirow{2}{*}{B} & $A \to B$ & 1 \\ 
                                  & & $B \to A$ & 0 \\
            \midrule
            \multirow{9}{*}{Confounding}  
               & \multirow{3}{*}{A} & $A \to B$ & 0 \\ 
                                 &  & $A \to C$ & 0 \\
                                 &  & $B \to C$ & 0  \\
                \cmidrule{2-4}
               & \multirow{3}{*}{B} & $A \to B$ & 1 \\ 
                                 &  & $A \to C$ & 0 \\
                                 &  & $B \to C$ & 0  \\
                \cmidrule{2-4}
               & \multirow{3}{*}{C} & $A \to B$ & 0 \\
                                 &  & $A \to C$ & 1 \\
                                 &  & $B \to C$ & 0 \\
            \midrule
            \multirow{9}{*}{Mediation}  
               & \multirow{3}{*}{A} & $A \to B$ & 0 \\
                                 &  & $A \to C$ & 0 \\
                                 &  & $B \to C$ & 0 \\
                \cmidrule{2-4}
               & \multirow{3}{*}{B} & $A \to B$ & 1 \\ 
                                 &  & $A \to C$ & 1 \\
                                 &  & $B \to C$ & 0 \\
                \cmidrule{2-4}
               & \multirow{3}{*}{C} & $A \to B$ & 0 \\
                                 &  & $A \to C$ & 1  \\
                                 &  & $B \to C$ & 1  \\
            \bottomrule
        \end{tabular}
\end{table}

We provide the ground truth value for every intervention effect task in our analysis in Table~\ref{tab:true_ie}, i.e., we provide the true change in the causal relations after interventions. 

This should help the reader understand clearly what queries we considered for the different research questions in our analysis. 





\clearpage

\section{Additional Results}
\label{app:extra-results}

We now go over the empirical studies in detail and provide the results for the remaining research questions.

\subsection{Reproducing results on the \tubingen dataset}

\label{sec:retreival-Tübingen}
We reproduced the results of \cite{kiciman2024causal} on the \tubingen dataset as shown in \Cref{tab:amit_tubingen}. Since their task did not involve the LLM to reason about the effect of interventions, we term this task of inferring relations from the input graph in the prompt as relation retrieval.

\begin{table}[!htbp]
    \caption{Relation Retrieval accuracy of GPT models on the \textbf{\tubingen} dataset. }
    \label{tab:amit_tubingen}
   \centering
    \begin{tabular}{ccc}
        \toprule
        \multicolumn{1}{c}{\bf Model} &\multicolumn{1}{c}{\bf Accuracy} \\
        \midrule
            GPT-3 (\textit{text-davinci-003}) & 0.80 \\ 
           GPT-3.5 & 0.89\\ 
           GPT-4 & 0.96\\ 
        \bottomrule
    \end{tabular}
\end{table}

\begin{table}[!htbp]
    \caption{\textbf{Relation Retrieval accuracy of models on Random benchmark.} The GPT models show good performance but LLaMA-2 performs poorly. Hence, as per our criteria we drop the LLaMA-2 model for analysis in RQ3.}
    \label{tab:retrieval_random_full}
    \centering
    \begin{tabular}{cccc}
        \toprule
        \multicolumn{1}{c}{\bf Model} &\multicolumn{1}{c}{\bf Bivariate} &\multicolumn{1}{c}{\bf Confounding} &\multicolumn{1}{c}{\bf Mediation} \\
        \midrule
            GPT-3.5 & $1.0 \pm 0.0$ & $1.0 \pm 0.0$ & $0.98 \pm 0.02$ \\
            GPT-4 & $1.0 \pm 0.0$ & $1.0 \pm 0.0$ & $1.0 \pm 0.0$ \\
            GPT-4-turbo & $1.0 \pm 0.0$ & $1.0 \pm 0.0$ & $1.0 \pm 0.0$ \\
            GPT-4o & $1.0 \pm 0.0$ & $1.0 \pm 0.0$ & $1.0 \pm 0.0$ \\
            LLaMA-2 & $0.73 \pm 0.11$ & $0.59 \pm 0.13$ & $0.84 \pm 0.09$ \\
            LLaMA-3 & $1.0 \pm 0.0$ & $1.0 \pm 0.0$ & $0.94 \pm 0.06$ \\
        \bottomrule
    \end{tabular}
\end{table}

\subsection{Research Questions}

\parhead{RQ1: How accurate are LLMs at predicting the effects of interventions?}
\label{app:extra-res-rq1}

To understand how effective LLMs are at IE prediction, we focused on the \textbf{Random (R)} benchmark in the main paper (\Cref{tab:ie_random}), as we wanted to remove any distracting effect of  semantically meaningful entities as graph nodes. We now present the same results on the \textbf{  \tubingen (T) } and \textbf{ Random \tubingen (RT) } benchmarks as well in \Cref{tab:ie_random_full}.

We find that the results in both the cases are very similar to the case with the \textbf{Random} benchmark; GPT-4-turbo performs the best and GPT models outperform LLaMA models by a big margin.

\begin{table}[h!]
\label{tab:variables}
     \caption{\textbf{IE prediction accuracy on all three benchmarks}. GPT-4 variants are the best performing models, while LLaMA models struggle with interventional reasoning.}
        \label{tab:ie_random_full}
    \centering
    \resizebox{\textwidth}{!}{
    \begin{tabular}{cccccccccc}
        \toprule
        Graph Type & Benchmark & \multicolumn{2}{c}{\bf Bivariate} &  \multicolumn{3}{c}{\bf Confounding} &  \multicolumn{3}{c}{\bf  Mediation} \\
        \midrule
        Intervened Variable & & A & B & A & B & C & A & B & C \\
        \midrule
        \multirow[c]{3}{*}{GPT-3.5} & R & $0.83 \pm 0.08$ & $0.87 \pm 0.06$ & $0.8 \pm 0.09$ & $0.69 \pm 0.12$ & $0.36 \pm 0.09$ & $0.58 \pm 0.11$ & $0.36 \pm 0.12$ & $0.67 \pm 0.12$ \\
        & T & $0.87 \pm 0.06$ & $0.83 \pm 0.06$ & $0.82 \pm 0.05$ & $0.67 \pm 0.11$ & $0.31 \pm 0.07$ & $0.64 \pm 0.08$ & $0.42 \pm 0.1$ & $0.67 \pm 0.12$ \\
        & RT & $0.83 \pm 0.1$ & $0.7 \pm 0.12$ & $0.87 \pm 0.09$ & $0.67 \pm 0.12$ & $0.47 \pm 0.13$ & $0.64 \pm 0.12$ & $0.33 \pm 0.12$ & $0.63 \pm 0.12$ \\
        \cmidrule{1-10}
       \multirow[c]{3}{*}{GPT-4} & R & $1.0 \pm 0.0$ & $1.0 \pm 0.0$ & $1.0 \pm 0.0$ & $1.0 \pm 0.0$ & $1.0 \pm 0.0$ & $0.78 \pm 0.09$ & $0.82 \pm 0.08$ & $0.96 \pm 0.03$ \\
       & T & $1.0 \pm 0.0$ & $1.0 \pm 0.0$ & $1.0 \pm 0.0$ & $1.0 \pm 0.0$ & $1.0 \pm 0.0$ & $1.0 \pm 0.0$ & $0.84 \pm 0.07$ & $1.0 \pm 0.0$ \\
        & RT & $1.0 \pm 0.0$ & $1.0 \pm 0.0$ & $0.93 \pm 0.06$ & $0.87 \pm 0.09$ & $1.0 \pm 0.0$ & $0.98 \pm 0.04$ & $0.97 \pm 0.05$ & $0.97 \pm 0.05$ \\
        \cmidrule{1-10}
        \multirow[c]{3}{*}{GPT-4-turbo} & R & $1.0 \pm 0.0$ & $0.97 \pm 0.03$ & $0.96 \pm 0.03$ & $0.93 \pm 0.05$ & $1.0 \pm 0.0$ & $0.98 \pm 0.02$ & $1.0 \pm 0.0$ & $1.0 \pm 0.0$ \\
        & T & $0.93 \pm 0.04$ & $0.9 \pm 0.05$ & $0.62 \pm 0.09$ & $0.78 \pm 0.08$ & $0.96 \pm 0.03$ & $0.69 \pm 0.09$ & $0.96 \pm 0.04$ & $0.96 \pm 0.04$ \\
        & RT & $0.8 \pm 0.1$ & $0.8 \pm 0.1$ & $0.73 \pm 0.11$ & $0.6 \pm 0.13$ & $1.0 \pm 0.0$ & $0.8 \pm 0.1$ & $0.97 \pm 0.05$ & $0.97 \pm 0.05$ \\
        \cmidrule{1-10}
        \multirow[c]{3}{*}{GPT-4o} & R & $1.0 \pm 0.0$ & $1.0 \pm 0.0$ & $0.98 \pm 0.02$ & $0.91 \pm 0.04$ & $1.0 \pm 0.0$ & $0.93 \pm 0.05$ & $1.0 \pm 0.0$ & $0.87 \pm 0.06$ \\
        & T & $0.87 \pm 0.06$ & $0.97 \pm 0.03$ & $0.89 \pm 0.04$ & $0.93 \pm 0.03$ & $1.0 \pm 0.0$ & $0.84 \pm 0.05$ & $0.96 \pm 0.04$ & $0.87 \pm 0.06$ \\
        & RT & $0.87 \pm 0.09$ & $0.83 \pm 0.1$ & $0.9 \pm 0.08$ & $0.8 \pm 0.1$ & $1.0 \pm 0.0$ & $0.93 \pm 0.06$ & $0.93 \pm 0.06$ & $0.87 \pm 0.09$ \\
         \cmidrule{1-10}
        \multirow[c]{3}{*}{LLaMA-2} & R & $0.5 \pm 0.12$ & $0.4 \pm 0.12$ & $0.56 \pm 0.11$ & $0.53 \pm 0.11$ & $0.16 \pm 0.06$ & $0.69 \pm 0.09$ & $0.56 \pm 0.12$ & $0.64 \pm 0.12$ \\
        & T & $0.4 \pm 0.12$ & $0.2 \pm 0.08$ & $0.42 \pm 0.11$ & $0.4 \pm 0.11$ & $0.09 \pm 0.04$ & $0.47 \pm 0.12$ & $0.31 \pm 0.12$ & $0.44 \pm 0.1$ \\
        & RT & $0.43 \pm 0.13$ & $0.23 \pm 0.11$ & $0.47 \pm 0.13$ & $0.47 \pm 0.13$ & $0.2 \pm 0.1$ & $0.62 \pm 0.13$ & $0.4 \pm 0.13$ & $0.7 \pm 0.12$ \\
         \cmidrule{1-10}
        \multirow[c]{3}{*}{LLaMA-3} & R & $0.8 \pm 0.1$ & $0.83 \pm 0.1$ & $0.47 \pm 0.13$ & $0.6 \pm 0.13$ & $0.87 \pm 0.09$ & $0.38 \pm 0.13$ & $0.5 \pm 0.13$ & $0.47 \pm 0.13$ \\
        & T & $0.73 \pm 0.11$ & $0.87 \pm 0.09$ & $0.53 \pm 0.13$ & $0.53 \pm 0.13$ & $1.0 \pm 0.0$ & $0.38 \pm 0.13$ & $0.6 \pm 0.13$ & $0.27 \pm 0.11$ \\
        & RT & $0.67 \pm 0.12$ & $0.87 \pm 0.09$ & $0.47 \pm 0.13$ & $0.4 \pm 0.13$ & $1.0 \pm 0.0$ & $0.4 \pm 0.13$ & $0.53 \pm 0.13$ & $0.43 \pm 0.13$ \\
        \bottomrule
    \end{tabular}}
\end{table}

\parhead{RQ2: To what extent is LLM performance affected by memorized causal relations?}
\label{app:extra-res-rq2} In \textbf{RQ1}, we computed the IE performace by considering all the causal relationship queries for each of the interventions. In the case of \textbf{RQ2}, we only consider specific causal relations that could be potentially memorized by the LLMs and interventions that sever these relations. For:
\begin{enumerate}
    \item Bivariate DAG, we consider $A \rightarrow B$ with the intervention on $B$.
    \item Confounding DAG, we consider $B \rightarrow C$ with the intervention on $C$. 
    \item Mediation DAG, we consider $A \rightarrow C$ with an intervention on $B$ and $C$ separately.
\end{enumerate}
The \textbf{\tubingen} benchmark (\Cref{app:dataset}) is defined in such a way that, all of the causal relations under consideration above are cause-effect pairs present in the TP dataset that could be potentially memorized by the LLM.

We provide results for role of memorization in LLaMA models in \Cref{tab:ie_memorization_full}. Given their overall poor IE performance in the random benchmark, it is difficult to draw conclusions about the impact of memorization but they have similar performance across the benchmarks.

\begin{table}
  \caption{\textbf{IE prediction performance of LLaMA models on specific scenarios for all the benchmarks to understand the role memorization.} LLaMA models perform poorly but demonstrate relatively similar performance across the three benchmarks.}
  \label{tab:ie_memorization_full}
  \centering
  \resizebox{\textwidth}{!}{
    \begin{tabular}{cccccc}
        \toprule 
        \textbf{Model} & \textbf{Graph} & \textbf{Intervened Variable} & \textbf{Random} & \textbf{\tubingen} & \textbf{Random \tubingen}  \\
        \midrule
    \multirow[c]{4}{*}{LlaMA-2} & \multirow[c]{1}{*}{Bivariate} & B  & $0.4 \pm 0.13$ & $0.4 \pm 0.13$ & $0.27 \pm 0.11$ \\
    \cmidrule{2-6} 
    & \multirow[c]{1}{*}{Confounding} & C & $0.0 \pm 0.0$ & $0.07 \pm 0.06$ & $0.33 \pm 0.12$ \\
     \cmidrule{2-6}
    & \multirow[c]{2}{*}{Mediation}  & B & $0.33 \pm 0.12$ & $0.33 \pm 0.12$  & $0.4 \pm 0.13$ \\
    &  & C & $0.53 \pm 0.13$ & $0.53 \pm 0.13$  & $0.6 \pm 0.13$ \\
    \cmidrule{1-6}
    \multirow[c]{4}{*}{LlaMA-3} & \multirow[c]{1}{*}{Bivariate} & B  & $0.67 \pm 0.12$ & $0.73 \pm 0.11$ & $0.8 \pm 0.1$ \\
    \cmidrule{2-6} 
    & \multirow[c]{1}{*}{Confounding} & C & $1.0 \pm 0.0$ & $1.0 \pm 0.0$ & $1.0 \pm 0.0$ \\
     \cmidrule{2-6}
    & \multirow[c]{2}{*}{Mediation}  & B & $0.67 \pm 0.12$ & $0.53 \pm 0.13$  & $0.67 \pm 0.12$ \\
    &  & C & $0.53 \pm 0.13$ & $0.33 \pm 0.12$ & $0.4 \pm 0.13$ \\
    \bottomrule
    \end{tabular}}
\end{table}

 \parhead{RQ3: Could LLMs be learning a shortcut to predict intervention effects?}
Consider the confounding DAG (\Cref{fig:graphs}) and the causal relation $C_{BC}(G)$ which does not change under intervention on the variable $A$. In these examples, LLMs that accurately parse causal relations from text descriptions of DAGs would also obtain accurate IE estimates. Hence, predicting causal relations from the input graph in text -- call this task relation retrieval -- offers a \emph{shortcut}: an LLM can attend to tokens in the context to solve relation retrieval and still perform good at IE estimation, thereby confounding the conclusions that can be drawn with this benchmark.

\begin{table}[]
   \caption{\textbf{IE prediction performance across sub-cases to isolate the effect of relation retrieval on Random benchmark}. LLMs do not significantly rely on shortcuts related to relation retrieval from the input prompt. We see a significant difference only for the confounding DAG case in GPT-3.5 and for bivariate DAG case in LLaMA-3. Since intervening on variable A never changes the causal graph in any scenario, we don't consider them for this analysis. Bolded figures indicate performances that are significantly ($\alpha=0.05$) different worse in IE = 1 compared to  IE = 0.}
    \label{tab:ie_change}
    \centering
    \resizebox{\textwidth}{!}{
        \begin{tabular}{ccccccc}
        \toprule
        Graph Type & &\multicolumn{1}{c}{\bf Bivariate} &  \multicolumn{2}{c}{\bf Confounding} &  \multicolumn{2}{c}{\bf  Mediation} \\
        \midrule
        Intervened Variable &  & B & B & C & B & C \\
        \midrule
        \multirow[c]{2}{*}{GPT-3.5} & IE = 0 & $0.93 \pm 0.06$ & $0.6 \pm 0.13$ & $\bf{0.47 \pm 0.13}$ & $0.33 \pm 0.12$ & $0.67 \pm 0.12$ \\
        \cmidrule{2-7}
        & IE = 1 & $0.8 \pm 0.1$ & $0.67 \pm 0.12$ & $\mb{0.2 \pm 0.1}$ & $0.37 \pm 0.12$ & $0.67 \pm 0.12$ \\
         \midrule
        \multirow[c]{2}{*}{GPT-4} & IE = 0 & $1.0 \pm 0.0$ & $1.0 \pm 0.0$ & $1.0 \pm 0.0$ & $0.87 \pm 0.09$ & $1.0 \pm 0.0$ \\
        \cmidrule{2-7}
        & IE = 1 & $1.0 \pm 0.0$ & $1.0 \pm 0.0$ & $1.0 \pm 0.0$ & $0.8 \pm 0.1$ & $0.93 \pm 0.06$ \\
        \midrule
        \multirow[c]{2}{*}{GPT-4-turbo} & IE = 0 & $1.0 \pm 0.0$ & $0.96 \pm 0.05$ & $1.0 \pm 0.0$ & $1.0 \pm 0.0$ & $1.0 \pm 0.0$ \\
        \cmidrule{2-7}
        & IE = 1  & $0.93 \pm 0.06$ & $0.93 \pm 0.06$ & $1.0 \pm 0.0$ & $1.0 \pm 0.0$ & $1.0 \pm 0.0$ \\
        \midrule
        \multirow[c]{2}{*}{GPT-4o} & IE = 0 & $1.0 \pm 0.0$ & $0.87 \pm 0.09$ & $1.0 \pm 0.0$ & $1.0 \pm 0.0$ & $0.87 \pm 0.09$ \\
        \cmidrule{2-7}
        & IE = 1  & $1.0 \pm 0.0$ & $1.0 \pm 0.0$ & $1.0 \pm 0.0$ & $1.0 \pm 0.0$ & $0.87 \pm 0.09$ \\
        \midrule
         \multirow[c]{2}{*}{LlaMA-2} & IE = 0 & $0.4 \pm 0.13$ & $0.5 \pm 0.13$ & $0.0 \pm 0.0$ & $0.33 \pm 0.12$ & $0.67 \pm 0.12$ \\
        \cmidrule{2-7}
        & IE = 1  & $0.4 \pm 0.13$ & $0.53 \pm 0.13$ & $0.13 \pm 0.09$ & $0.33 \pm 0.12$ & $0.6 \pm 0.13$ \\
        \midrule
        \multirow[c]{2}{*}{LlaMA-3} & IE = 0 & $\bf{1.0 \pm 0.0}$ & $0.4 \pm 0.13$ & $1.0 \pm 0.0$ & $0.6 \pm 0.13$ & $0.27 \pm 0.11$ \\
        \cmidrule{2-7}
        & IE = 1  & $\bf{0.67 \pm 0.12}$ & $0.6 \pm 0.13$ & $0.87 \pm 0.09$ & $0.5 \pm 0.13$ & $0.47 \pm 0.13$ \\
        \bottomrule
    \end{tabular}}
    \label{tab:my_label}
\end{table}

However, the intervention effects we defined to study RQ2 offer an insight into how we can disentangle relation retrieval and accurate IE prediction. Notice that the IEs we defined to study RQ2 characterize scenarios where causal relations differ between the base DAG $G$ and the post-intervention DAG. Thus, LLMs that rely only on relation retrieval cannot accurately estimate IEs. Building on this insight, we divide all the IEs $\kappa^i_G(C_{uv})$ into two groups:
\begin{enumerate}
    \item {$\bf IE= 0$}: Graph doesn't change as a result of intervention; $C_{uv}(G) = C_{uv}(G^i)$.
    \item {$\bf IE= 1$}: Graph changes as a result of intervention; $C_{uv}(G) \neq C_{uv}(G^i)$
\end{enumerate} 

\Cref{tab:true_ie} clearly categorizes the causal relations under intervention into these two groups.

Note that drop in performance on group ($IE=1$) as compared to the group ($IE=0$) can indicate reliance on shortcuts based on relation retrieval. We consider the average performance of the LLMs on the effects in each group, selecting only those LLMs which achieve an accuracy $\geq 0.95$ on relation retrieval (which we report in Appendix \Cref{tab:retrieval_random_full}). We focus on the \textbf{Random} benchmark to exclude any impacts due to memorized variable names. \Cref{tab:ie_change} summarizes this study. We find that the general trend does not show strong reliance on shortcuts across LLMs;  GPT-3.5 only has a significant drop in its relative performance on group ($IE=1$) vs group ($IE=0$) for the confounding DAG cases. And, significant difference for bivariate DAG cases in LLaMA-3.  Since, LLaMA-2 doesn't satisfy our constraint of high relation retrieval accuracy, we don't consider it for further analysis.

\parhead{RQ4: Are LLMs robust to descriptions of interventions in-context?}
Consider the verbalization of an intervention in \Cref{fig:prompt}. It mentions the $\rmdo(\cdot)$ operator and refers to a ``perfect intervention'' to prompt an LLM to rely on facts about causal reasoning that could have appeared in the training dataset.
We ask whether LLMs can achieve the same performance on the intervention effects in the random benchmark if we varied the verbalization to instead ``teach'' an LLM~\citep{patel2023magnifico} about a new graphical operation that behaves identically to an intervention. We randomly generate strings to instantiate this operation. \Cref{fig:prompt-substitution} in the appendix illustrates how we generate prompts for this task which we refer to as the substitution task.
\Cref{tab:ie_substitution} summarizes IE prediction accuracy for the substitution task on the \textbf{Random} benchmark. Contrasting these results against \Cref{tab:ie_random}, we see that for the GPT-3.5, GPT-4o, and LLaMA models, performance generally suffers. However, GPT-4-turbo appears to be robust to changes in the way interventions are described. 

\begin{table}[]
    \caption{\textbf{IE prediction accuracy on the Random benchmark for the substitution task}. Except for GPT-4 and GPT-4-turbo, rest of the models perform worse under substitution compared to the non-substitution case (Table~\ref{tab:ie_random}).}
    \label{tab:ie_substitution}
    \centering
    \resizebox{\textwidth}{!}{
    \begin{tabular}{ccccccccc}
        \toprule
        Graph Type &  \multicolumn{2}{c}{\bf Bivariate} &  \multicolumn{3}{c}{\bf Confounding} &  \multicolumn{3}{c}{\bf  Mediation} \\
        \midrule
        Intervened Variable & A & B & A & B & C & A & B & C \\
        \midrule
        GPT-3.5 & $0.67 \pm 0.12$ & $0.83 \pm 0.06$ & $0.56 \pm 0.11$ & $0.58 \pm 0.11$ & $1.0 \pm 0.0$ & $0.62 \pm 0.10$ & $0.44 \pm 0.10$ & $0.4 \pm 0.12$ \\
        \cmidrule{1-9}
        GPT-4 & $0.97 \pm 0.03$ & $1.0 \pm 0.0$ & $0.89 \pm 0.05$ & $0.82 \pm 0.07$ & $1.0 \pm 0.0$ & $0.96 \pm 0.03$ & $0.76 \pm 0.10$ & $1.0 \pm 0.0$ \\
        \cmidrule{1-9}
        GPT-4-turbo & $0.9 \pm 0.07$ & $0.97 \pm 0.03$ & $0.96 \pm 0.03$ & $1.0 \pm 0.0$ & $1.0 \pm 0.0$ & $1.0 \pm 0.0$ & $1.0 \pm 0.0$ & $1.0 \pm 0.0$ \\
        \cmidrule{1-9}
        GPT-4o & $0.67 \pm 0.12$ & $0.83 \pm 0.1$ & $0.67 \pm 0.12$ & $0.73 \pm 0.11$ & $1.0 \pm 0.0$ & $0.76 \pm 0.11$ & $0.83 \pm 0.1$ & $0.73 \pm 0.11$ \\
        \cmidrule{1-9}
        LLaMA-2 & $0.57 \pm 0.13$ & $0.37 \pm 0.12$ & $0.6 \pm 0.13$ & $0.6 \pm 0.13$ & $0.73 \pm 0.11$ & $0.62 \pm 0.13$ & $0.6 \pm 0.13$ & $0.43 \pm 0.13$ \\
        \cmidrule{1-9}
        LLaMA-3 & $0.67 \pm 0.12$ & $0.83 \pm 0.1$ & $0.6 \pm 0.13$ & $0.4 \pm 0.13$ & $1.0 \pm 0.0$ & $0.4 \pm 0.13$ & $0.67 \pm 0.12$ & $0.33 \pm 0.12$ \\
        \bottomrule
    \end{tabular}}
\end{table}

\end{document}